\newcommand{\CLASSINPUTinnersidemargin}{18mm}
\newcommand{\CLASSINPUToutersidemargin}{12mm}
\newcommand{\CLASSINPUTtoptextmargin}{20mm}
\newcommand{\CLASSINPUTbottomtextmargin}{25mm}
\documentclass[10pt,conference,a4paper]{IEEEtran}

\usepackage[utf8]{inputenc}

\usepackage{times}

\usepackage{graphicx}
\usepackage{subfigure}

\usepackage{multirow}

\DeclareGraphicsExtensions{.png,.eps,.ps,.pdf}

\usepackage{url}

\usepackage[english]{babel}

\begin{document}

\title{Perceptual Hashing applied to Tor domains recognition}

\makeatletter
\newcommand{\linebreakand}{%
  \end{@IEEEauthorhalign}
  \hfill\mbox{}\par
  \mbox{}\hfill\begin{@IEEEauthorhalign}
}
\makeatother

\author{
\IEEEauthorblockN{\small Rubel Biswas}
\IEEEauthorblockA{\small
Dept. IESA.\\
Universidad de León\\
Researcher at INCIBE\\
rbis@unileon.es}

\and

\IEEEauthorblockN{\small Roberto A. Vasco-Carofilis}
\IEEEauthorblockA{\small
Dept. IESA.\\
Universidad de León\\
Researcher at INCIBE\\
rvasc@unileon.es}

\and 

\IEEEauthorblockN{\small Eduardo Fidalgo}
\IEEEauthorblockA{\small
Dept. IESA.\\
Universidad de León\\
Researcher at INCIBE\\
eduardo.fidalgo@unileon.es}

\and

\IEEEauthorblockN{\small Francisco Jañez-Martino}
\IEEEauthorblockA{\small
Dept. IESA.\\
Universidad de León\\
Researcher at INCIBE\\
fjanm@unileon.es}

\and

\IEEEauthorblockN{\small Pablo Blanco-Medina}
\IEEEauthorblockA{\small
Dept. IESA.\\
Universidad de León\\
Researcher at INCIBE\\
pblanm@unileon.es}}
\maketitle

\begin{abstract}

The Tor darknet hosts different types of illegal content, which are monitored by cybersecurity agencies. However, manually classifying Tor content can be slow and error-prone. To support this task, we introduce Frequency-Dominant Neighborhood Structure (F-DNS), a new perceptual hashing method for automatically classifying domains by their screenshots. First, we evaluated F-DNS using images subject to various content-preserving operations. We compared them with their original images, achieving better correlation coefficients than other state-of-the-art methods, especially in the case of rotation. Then, we applied F-DNS to categorize Tor domains using the Darknet Usage Service Images-2K (DUSI-2K), a dataset with screenshots of active Tor service domains. Finally, we measured the performance of F-DNS against an image classification approach and a state-of-the-art hashing method. Our proposal obtained 98.75\% accuracy in Tor images, surpassing all other methods compared.
\end{abstract}


{\bf Type of contribution:}  {\it Research already published}

\section{Introduction}


The Deep Web content cannot be indexed by standard search engines, such as Google, Yahoo, or Bing~\cite{fidalgo2019classifying}. Within it, we find darknets that can only be accessed by unique browsers such as Tor (The Onion Router). These domains host various kinds of suspicious content \cite{gangwar2017pornography}.

According to Al-Nabki et al., at least $20\%$ of the content found in Tor domains can be considered as illegal~\cite{Nabki2019ToRankIT}, so Law Enforcement Agencies (LEA) are interested in monitor Tor darknets \cite{fidalgo2019classifying, al2017detecting, FidalgoRIAI2019}. The manual categorization of the Darknet is not feasible due to the amount of data availability, requiring the use of automatic tools to identify and classify Tor darknet domains.

To support this task, we present and make publicly available Darknet Usage Service Images-2K (DUSI-2K)\footnote{\url{ http://gvis.unileon.es/dataset/dusi-darknet-usage-service-images-2k/}}, a dataset with $2500$ snapshots of Tor domain home pages, divided into $16$ categories.

Furthermore we introduce Frequency-Dominant Neighbourhood Structure (F-DNS), a new perceptual hashing method that demonstrates excellent performance against image content-preserving operations, like scaling. Finally, we applied F-DNS to the problem of classifying Tor domains using its screenshots and compare its performance with other state-of-the-art methods~\cite{BISWAS202024}.

\section{Darknet Usage Service Images-2K (DUSI-2K)}

DUSI-2K dataset is built in a semi-supervised way, extending the Darknet Usage Service Images (DUSI) \cite{Rubel2017SIR} dataset by including snapshots from 16 classes of active Tor domains. All domains were crawled using the labeled domains of Darknet Usage Text Addresses (DUTA) dataset~\cite{Nabki_17ClassIllegalTORText}.


\section{Construction of F-DNS hash}

The pipeline of our F-DNS method is presented in Fig.~\ref{fig:pipeline}. In pre-processing, the image is converted to grayscale and smoothed using a Gaussian filter.

\begin{figure}[t]
\centerline{
\includegraphics[width=0.9\linewidth]{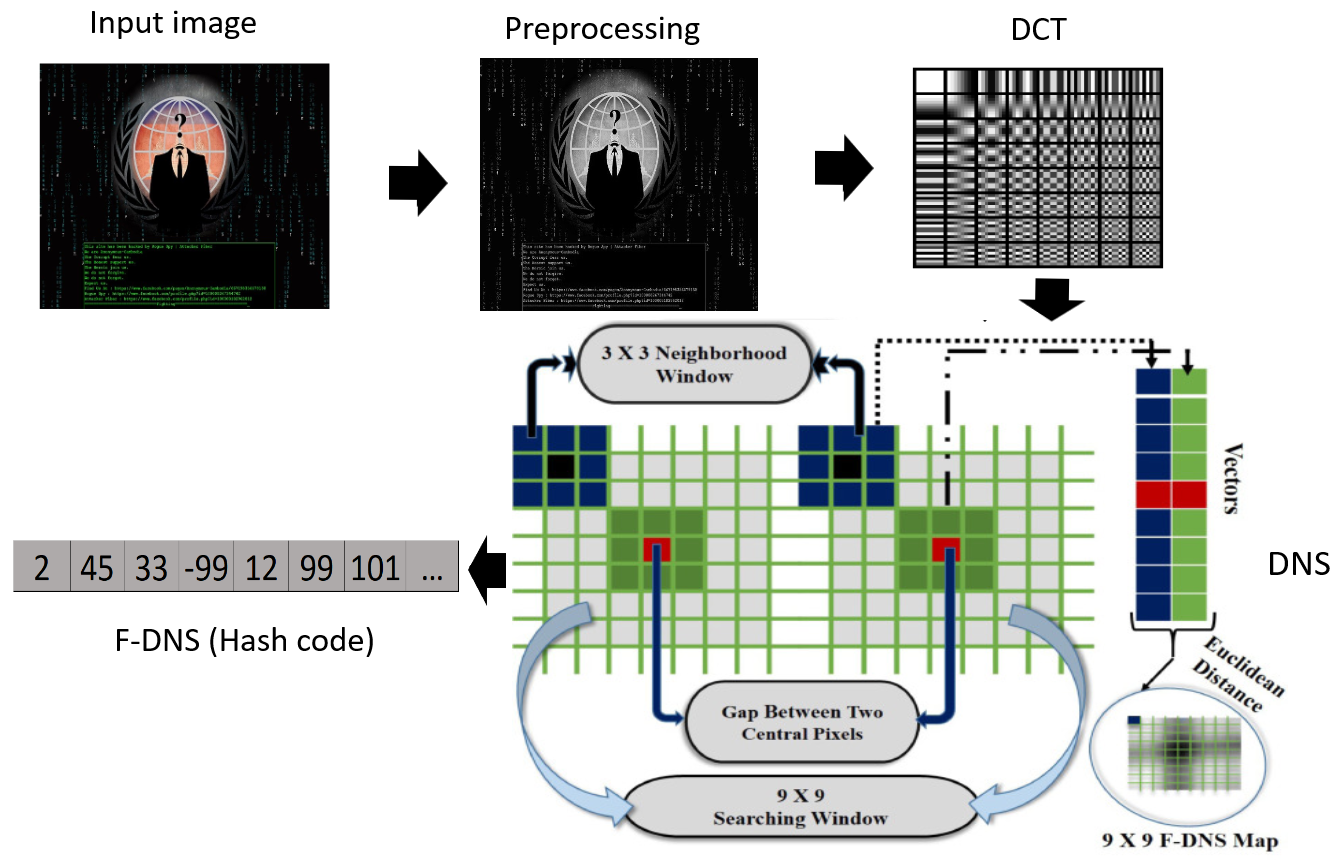}
}
\caption{Pipeline of F-DNS hashing method.}
\label{fig:pipeline}
\end{figure}

After pre-processing, the image features are extracted employing Discrete Cosine Transform (DCT)~\cite{ahmed1974discrete} and the Dominant Neighborhood Structure (DNS), proposed by Khellah~\cite{Khellah2011TextureClassDNS}. Since the DNS is extracted from the DCT of the image, we named the extracted map as Frequency-Dominant Neighborhood Structure (F-DNS).

First, we apply DCT to the pre-processed image and then the DNS \cite{Khellah2011TextureClassDNS} is applied on the output of the DCT of the image to extract features from its texture energies. The DNS exploits the high redundancy that is found on images with repetitive patterns. 

Given a pixel $x$, called \textit{central pixel}, the DNS, $D$, is obtained by computing the intensity similarity for all pixels $x^{\prime}$ which fall within a $N \times N$ neighborhood around it, called \textit{searching window}. The similarity of each pixel $x^{\prime}$ of the searching window is calculated employing the Euclidean distance between the intensities in the flattened matrices of $M \times M$ pixels around both $x$ and $x^{\prime}$. This area of $M \times M$ pixels is called \textit{neighborhood window}. 
If the coordinates of $x^{\prime}$ within the neighborhood window are $(i,j)$, then the similarity between $x$ and $x^{\prime}$ is placed in the position $(i,j)$ of the DNS, i.e. $D(i,j)$. Therefore, the DNS represents the degree of similarity of texture energies between pixel $x$ and its neighbor pixels. 



After obtaining $N$ F-DNS maps, we compute the Frequency-Global Neighborhood Structure (F-GNS) of the image by summing up all F-DNS maps from the image.
\par

The final image hash is obtained using the coefficients but discarding the first row and column, to avoid including the average of the pixel values, obtained during the DCT calculation process. At the end of the process, the hash code of each image is composed of $64$ real values.

In this work, we have considered $9 \times 9$ pixels (i.e. $N=9$) searching window and $3 \times 3$ pixels neighborhood window. 

\section{Experimental results}
To evaluate the robustness of F-DNS, we used USC-SIPI~\cite{sipi2016usc} state-of-the-art dataset to generate visually identical versions of $35$ images, applying various content-preserving operations. We calculated the correlation coefficients between the hashes obtained from the altered images and the hash of their corresponding original image.

We compared the performance of F-DNS against RP-IVD (Ring Partition and Invariant Vector Distance)~\cite{Tang1_16RobImaHashRingPartition}, a state-of-the-art perceptual hashing method. The average score of the correlation coefficients obtained in each task can be seen in Table~\ref{tab:USCMeanScore}.


\begin{table}[!htb]
\caption{Mean correlation coefficient scores of each method.}
\begin{center}
\resizebox{\linewidth}{!} {
\begin{tabular}{c c c c c c}
\hline
\hline
\textbf{Operation} & \textbf{RP-IVD} & \textbf{F-DNS} & \textbf{Operation} & \textbf{RP-IVD} & \textbf{F-DNS}\\
\hline
Brightness adj. & 0.9583 & \textbf{0.9985} & Gaussian filter & 0.9973 & \textbf{0.9999}  \\
Contrast adj & 0.9920 & \textbf{0.9993} & JPEG compression & 0.9986 & \textbf{0.9993}   \\
Gamma correction & 0.9957 & \textbf{0.9995} & Scaling & 0.9773 & \textbf{0.9875}   \\
Salt \& pepper noise & 0.9872 & \textbf{0.9999} & Rotation & 0.2959& \textbf{0.9365}  \\
Multiplicative noise & 0.9939 & \textbf{0.9999} &  Watermark embedding & 0.9601 & \textbf{0.9989} \\

\hline
\hline
\end{tabular}
}
\end{center}
\label{tab:USCMeanScore}
\end{table}

Our proposal performs best against most content-preserving operations, and stands out for its performance in rotation, which is one of its major advantages over similar proposals.






Additionally, we tested our proposal using the Tor domain screenshots taken from the DUSI-2K dataset. We took a total of $1624$ images, from which we selected $79$ templates,  i.e. snapshots of domains, which are frequently used in Tor domains with similar topics. Therefore, by determining which template is most similar to each screenshot we can deduce which category each screenshot belongs to.

We calculated the hash codes of the templates and compared them with the hash code of each of the remaining $1545$ images. For classification, images are assigned labels based on the template with the highest similarity. We measured performance using the accuracy metric.

We repeated the experiment $20$ times, including the random selection of the template from the images in each class, and compared F-DNS against RP-IVD. Since this approach can be considered as an image classification task~\cite{fidalgo2018boosting}, we also reported the results with state-of-the-art image descriptors, such as Inception-ResNet-v2 \cite{Szegedy2017InceptionResNet}. We split split DUSI-2K randomly into $5$ disjoint sets, setting $75$\% of the images for training a Support Vector Machine (SVM) with linear kernel, and $25$\% for testing. The results are shown in Table~\ref{tab:ClassificationScore}.


\begin{table}[!htb]
\caption{Tor domain classification accuracy in DUSI-2K.}
\begin{center}
\begin{tabular}{c c}
\hline
\hline
\textbf{Methods} & \textbf{Overall accuracy}\\
\hline
RP-IVD & 95.84\%  \\
Inception-ResNet-v2 & 85.19\%  \\
F-DNS & \textbf{98.75\%}  \\
\hline
\hline
\end{tabular}
\end{center}
\label{tab:ClassificationScore}
\end{table}

\section{Conclusions}

In this paper, we presented DUSI-2K, a dataset with $2500$ snapshots of Tor domains. We have also proposed a new robust image hashing scheme, called F-DNS, and used it to classify Tor domains.

We compared the performance of F-DNS with other state-of-the-art hashing schemes, as well as image classification models, demonstrating that F-DNS achieved the best results.

\section*{Acknowledgements}

This work was supported by the framework agreement between the Universidad de Le\'{o}n and INCIBE (Spanish National Cybersecurity Institute) under Addendum $01$.
We acknowledge NVIDIA Corporation with the donation of the TITAN Xp and Tesla K40 GPUs used for this research.

\bibliographystyle{IEEEtran}
\bibliography{biblio}

\begin{thebibliography}{10}
\providecommand{\url}[1]{#1}
\csname url@samestyle\endcsname
\providecommand{\newblock}{\relax}
\providecommand{\bibinfo}[2]{#2}
\providecommand{\BIBentrySTDinterwordspacing}{\spaceskip=0pt\relax}
\providecommand{\BIBentryALTinterwordstretchfactor}{4}
\providecommand{\BIBentryALTinterwordspacing}{\spaceskip=\fontdimen2\font plus
\BIBentryALTinterwordstretchfactor\fontdimen3\font minus
  \fontdimen4\font\relax}
\providecommand{\BIBforeignlanguage}[2]{{%
\expandafter\ifx\csname l@#1\endcsname\relax
\typeout{** WARNING: IEEEtran.bst: No hyphenation pattern has been}%
\typeout{** loaded for the language `#1'. Using the pattern for}%
\typeout{** the default language instead.}%
\else
\language=\csname l@#1\endcsname
\fi
#2}}
\providecommand{\BIBdecl}{\relax}
\BIBdecl

\bibitem{fidalgo2019classifying}
E.~Fidalgo, E.~Alegre, L.~Fern{\'a}ndez-Robles, and V.~Gonz{\'a}lez-Castro,
  ``Classifying suspicious content in tor darknet through semantic attention
  keypoint filtering,'' \emph{Digital Investigation}, vol.~30, pp. 12--22,
  2019.

\bibitem{gangwar2017pornography}
A.~Gangwar, E.~Fidalgo, E.~Alegre, and V.~Gonz{\'a}lez-Castro, ``Pornography
  and child sexual abuse detection in image and video: A comparative
  evaluation,'' in \emph{Imaging for Crime Detection and Prevention}, 2017.

\bibitem{Nabki2019ToRankIT}
M.~W.~A. Nabki, E.~Fidalgo, E.~Alegre, and L.~Fern{\'a}ndez-Robles, ``{ToRank:
  Identifying the most influential suspicious domains in the Tor network},''
  \emph{{Expert Systems with Applications}}, vol. 123, pp. 212--226, 2019.

\bibitem{al2017detecting}
M.~W. Al-Nabki, E.~Fidalgo~Fern{\'a}ndez, E.~Alegre~Guti{\'e}rrez,
  V.~Gonz{\'a}lez~Castro \emph{et~al.}, \emph{Detecting emerging products in
  tor network based on k-shell graph decomposition}, 2017.

\bibitem{FidalgoRIAI2019}
E.~{Fidalgo Fern{\'{a}}ndez}, E.~{Alegre Guti{\'{e}}rrez}, L.~{Fern{\'{a}}ndez
  Robles}, and V.~{Gonz{\'{a}}lez Castro}, ``{Fusi{\'{o}}n temprana de
  descriptores extra{\'{i}}dos de mapas de prominencia multi-nivel para
  clasificar im{\'{a}}genes},'' \emph{Revista Iberoamericana de
  Autom{\'{a}}tica e Inform{\'{a}}tica.}, vol.~16, no.~3, pp. 358--368, 2019.

\bibitem{BISWAS202024}
R.~Biswas, V.~Gonz\'{a}lez-Castro, E.~Fidalgo, and E.~Alegre, ``Perceptual
  image hashing based on frequency dominant neighborhood structure applied to
  tor domains recognition,'' \emph{Neurocomputing}, vol. 383, pp. 24 -- 38,
  2020.

\bibitem{Rubel2017SIR}
R.~Biswas, E.~Fidalgo, and E.~Alegre, ``{Recognition of Service Domains on TOR
  Dark Net using Perceptual Hashing and Image Classification Techniques},'' in
  \emph{{8th International Conference on Imaging for Crime Detection and
  Prevention (ICDP)}}, I.~D. Library, Ed., 2017, pp. 13--15.

\bibitem{Nabki_17ClassIllegalTORText}
M.~W.~A. Nabki, E.~Fidalgo, E.~Alegre, and I.~de~Paz, ``{Classifying illegal
  activities on TOR network based on web textual contents},'' in
  \emph{{Proceedings of the 15th Conference of the European Chapter of the
  Association for Computational Linguistics}}, vol.~1, 2017, pp. 35--43.

\bibitem{ahmed1974discrete}
N.~Ahmed, T.~Natarajan, and K.~R. Rao, ``Discrete cosine transform,''
  \emph{IEEE transactions on Computers}, vol. 100, no.~1, pp. 90--93, 1974.

\bibitem{Khellah2011TextureClassDNS}
F.~M. Khellah, ``{Texture classification using dominant neighborhood
  structure},'' \emph{{IEEE Transactions on Image Processing}}, vol.~20,
  no.~11, pp. 3270--3279, 2011.

\bibitem{sipi2016usc}
USC-SIPI, ``The usc-sipi image database,'' \url{http://sipi.usc.edu/database/},
  2016.

\bibitem{Tang1_16RobImaHashRingPartition}
Z.~Tang, X.~Zhang, X.~Li, and S.~Zhang, ``{Robust image hashing with ring
  partition and invariant vector distance},'' \emph{{IEEE Transactions on
  Information Forensics and Security}}, vol.~11, no.~1, pp. 200--214, 2016.

\bibitem{fidalgo2018boosting}
E.~Fidalgo, E.~Alegre, V.~Gonzalez-Castro, and L.~Fern{\'a}ndez-Robles,
  ``Boosting image classification through semantic attention filtering
  strategies,'' \emph{Pattern Recognition Letters}, vol. 112, pp. 176--183,
  2018.

\bibitem{Szegedy2017InceptionResNet}
C.~Szegedy, S.~Ioffe, V.~Vanhoucke, and A.~A. Alemi, ``{Inception-v4,
  Inception-ResNet and the Impact of Residual Connections on Learning},'' in
  \emph{{Proceedings of the Thirty-First AAAI Conference on Artificial
  Intelligence}}, ser. AAAI'17, 2017, pp. 4278--4284.

\end{thebibliography}

\end{document}